\documentclass{article}
\usepackage{ijcai21}
\usepackage{comment}

\usepackage{times}

\usepackage{soul}
\usepackage{url}
\usepackage[hidelinks]{hyperref}
\usepackage[utf8]{inputenc}
\usepackage[small]{caption}
\usepackage{graphicx}
\usepackage{amsmath}
\usepackage{booktabs}
\makeatletter
\newcommand{\printfnsymbol}[1]{%
  \textsuperscript{\@fnsymbol{#1}}%
}
\makeatother

\title{Term Expansion and FinBERT fine-tuning for Hypernym and Synonym Ranking of Financial Terms}

\author{
Ankush Chopra\footnote{Contact Author}\thanks{Equal Contribution}\and
Sohom Ghosh\printfnsymbol{2}
\affiliations
Fidelity Investments, AI CoE, Bengaluru, India\\
\emails
\{ankush01729, sohom1ghosh\}@gmail.com
}

\begin{document}

\maketitle

\begin{abstract}
Hypernym and synonym matching are one of the mainstream Natural Language Processing (NLP) tasks. In this paper, we present systems that attempt to solve this problem. We designed these systems to participate in the FinSim-3, a shared task of FinNLP workshop at IJCAI-2021. The shared task is focused on solving this problem for the financial domain. We experimented with various transformer based pre-trained embeddings by fine-tuning these for either classification or phrase similarity tasks. We also augmented the provided dataset with abbreviations derived from prospectus provided by the organizers and definitions of the financial terms from DBpedia \cite{dbpedia}, Investopedia, and the Financial Industry Business Ontology (FIBO). Our best performing system uses both FinBERT \cite{araci2019finbert} and data augmentation from the afore-mentioned sources. We observed that term expansion using data augmentation in conjunction with semantic similarity is beneficial for this task and could be useful for the other tasks that deal with short phrases. Our best performing model (Accuracy: 0.917, Rank: 1.156) was developed by fine-tuning SentenceBERT \cite{sbert} (with FinBERT at the backend) over an extended labelled set created using the hierarchy of labels present in FIBO.
\end{abstract}

\section{Introduction}
Ontologies are rich sources of information that provide deep information about the underlying concepts and entities. This information is described for a specific domain. It contains the clearly defined relationships, and it is organized in a defined structure mostly as a hierarchy. These properties make ontologies a great source for getting a deeper understanding of the relationship and properties of resources of the domain in consideration.

Public knowledge graphs and ontologies like DBpedia and Yago have been shown to work on various applications like the ones described in \cite{dbpediabbc} and \cite{hahm2014named}. This has motivated and paved ways for the creation of domain focused ontologies like FIBO\footnote{https://spec.edmcouncil.org/fibo/}. 

Effective techniques that enable identifying lexical similarity between the terms or concepts increase the effectiveness of the ontologies. These methods not only help in building new ontologies faster or augment the existing ones, but also it helps in the effective querying and searching of concepts. 

FinSim \cite{maarouf-etal-2020-finsim,all-2021-finsim} competitions are being held to promote the development of effective similarity measures. In the third edition of the competition FinSim-3\footnote{https://sites.google.com/nlg.csie.ntu.edu.tw/finnlp2021/shared-task-finsim (accessed on 8\textsuperscript{th} July 2021)} (being held in conjunction with the 30\textsuperscript{th} International Joint Conference on Artificial Intelligence (IJCAI-21)), the participants are challenged to develop methods and systems to rank hypernym and synonyms to financial terms by mapping them to one of the 17 high-level financial concepts present in FIBO.

In this paper, we present the systems developed by our team Lipi for hypernym and synonym ranking. We experimented with basic featurization methods like TF-IDF and advanced methods like pre-trained embedding models. Our top 3 systems use pre-trained FinBERT \cite{araci2019finbert} embedding model that was fine-tuned on the data specific to financial domain . We also augmented the training data by utilizing the knowledge from DBpedia, Investopedia, FIBO and text corpus of prospectus shared with us.
We describe the works related to our solution in the next section. Section 3 contains the formal problem statement, followed by data description in section 4. We describe our top three systems in section 5. Section 6 contains the details of the experimentation that we performed and the results obtained from some of them. We draw our conclusions in section 7 while giving a glimpse of things that we would like to try in the future.

\section{Related Works}
Hypernym-hyponym extraction and learning text similarity using semantic representations have been very challenging areas of research for the NLP community. SemEval-2018 Task 9 \cite{camacho-collados-etal-2018-semeval} was such an instance. Team CRIM \cite{bernier-colborne-barriere-2018-crim} performed the best in this shared task. They combined a supervised word embedding based approach with an unsupervised pattern discovery based approach.  The FinSim shared tasks \cite{maarouf-etal-2020-finsim,all-2021-finsim} deal with adopting these challenges specific to the Financial Domain. Team IIT-K \cite{keswani-etal-2020-iitk} won FinSim-1 using a combination of context-free static embedding Word2Vec \cite{word2vec} and contextualized dynamic embedding BERT \cite{devlin-etal-2019-bert}. Anand et al. \cite{anand-etal-2020-finsim20} from the team FINSIM20 explored the use of cosine similarity between terms and labels encoded using Universal Sentence Encoder \cite{cer2018universal}. They also tried to extract hypernyms automatically using graph based approaches. Team PolyU-CBS \cite{pplyu-2021-finsim} won FinSim-2 shared task using Logistic Regression trained over word embedding and probabilities derived from BERT \cite{devlin-etal-2019-bert} model. They also experimented with GPT-2 \cite{gpt2}. Team L3i-LBPAM \cite{l3i-2021-finsim} comprising Nguyen et al. performed better than the baseline by using Sentence BERT \cite{sbert} to calculate cosine similarity between terms and hypernyms. \cite{saini-2020-anuj-investopedia,goat-2021-finsim} and \cite{jurgens-pilehvar-2016-semeval} discussed various techniques to enrich the data which was available for training. In this edition of FinSim, the number of training samples and labels (financial concepts) were more than the previous two editions.

\section{Problem Statement}
Given a set F consisting of n tuples of financial terms and their hypernyms/top-level concepts/labels i.e. $F = \{(t\textsubscript{1},h\textsubscript{1}), (t\textsubscript{2},h\textsubscript{2}), ... (t\textsubscript{n},h\textsubscript{n})\}$ where h\textsubscript{i} represents the hypernym corresponding to the i\textsuperscript{th} term t\textsubscript{i} and $h\textsubscript{i} \epsilon$ set of labels mentioned in Table \ref{tab:labeldis}. 
For every unseen financial term, our task is to generate a ranked list $\hat{y_i}$ consisting of these 17 hypernyms in order of decreasing semantic similarity.\\

\textbf{Evaluation Metrics}
The expected output is a raked list of predicted labels for every scored instance. The proposed systems are evaluated based on Accuracy and Mean Rank metrices as per the shared task rules. Evaluation script was provided by organizers, where accuracy and mean rank were defined as:\\
$Accuracy = \frac{1}{n}\sum_{i=1}^{n}I(y_i=\hat{y_i}[1])$\\
$Mean Rank = \frac{1}{n}\sum_{i=1}^{n}(\hat{y_i}.index(y_i))$\\
where $\hat{y_i}$ is the ranked list (with index starting from 1) of predicted labels corresponding to the expected label $y_i$. I is an identity matrix.
\section{Data}
\subsection{Data Description}
The training dataset shared for this task has a total of 1050 single and multi-word terms tagged to 17 different classes/labels out of which 1040 term-label pairs are unique. More than 91\% of the terms have 6 words or less and the longest term has 22 words. There were 10 duplicate entries, and 3 terms were assigned 2 different labels. Along with this, a corpus of prospectuses in English that had 211  documents was provided. Some of the terms mentioned in the training data were present in the corpus. Table \ref{tab:labeldis} shows the distribution of these labels in the training set.
\begin{table}
\centering
\begin{tabular}{|l|l|}
\hline
\textbf{Label}                & \textbf{Count} \\ \hline
Equity Index                  & 280            \\ \hline
Regulatory Agency             & 205            \\ \hline
Credit Index                  & 125            \\ \hline
Central Securities Depository & 107            \\ \hline
Debt pricing and yields       & 58             \\ \hline
Bonds                         & 55             \\ \hline
Swap                          & 36             \\ \hline
Stock Corporation             & 25             \\ \hline
Option                        & 24             \\ \hline
Funds                         & 22             \\ \hline
Future                        & 19             \\ \hline
Credit Events                 & 18             \\ \hline
MMIs                          & 17             \\ \hline
Stocks                        & 17             \\ \hline
Parametric schedules          & 15             \\ \hline
Forward                       & 9              \\ \hline
Securities restrictions       & 8              \\ \hline
\textbf{Total}                & \textbf{1040}  \\ \hline
\end{tabular}
\caption{Label distribution in the training set}
\label{tab:labeldis}
\end{table}

\subsection{Data Augmentation}
\label{sec:dataaug}
Since the majority of the terms had only a few tokens, we decided to expand the terms wherever possible using various sources. This approach had also been adopted by \cite{saini-2020-anuj-investopedia} and \cite{goat-2021-finsim} while participating in FinSim-1 and FinSim-2 respectively. \\

\textbf{Acronym expansion}: As mentioned by Keswani et al. \cite{keswani-etal-2020-iitk}, the presence of acronyms created a major issue in maintaining consistency.  We used the abbreviation extractor available in  spaCy\footnote{https://spacy.io/}\cite{spacy} package on the corpus of the prospectus to extract all the acronyms and their expansions. Upon manual inspection of a sample output, we identified that not all the extracted items were valid acronyms and their expansions. We cleaned the extracted list by dropping the records where:
\begin{itemize}
    \item expansion had equal or less length than the acronym.
    \item expansion had parenthesis
    \item extracted acronym was a valid English word such as "fund" or "Germany".
    \item the expansion had less than or equal to 5 characters.
\end{itemize}
We managed to extract 635 acronyms from the prospectus corpus after applying the above exclusions. We used this data to expand the matching terms in the given train set and test sets.\\

\begin{figure*}
\centering
  \includegraphics[width=\textwidth,height=4cm]{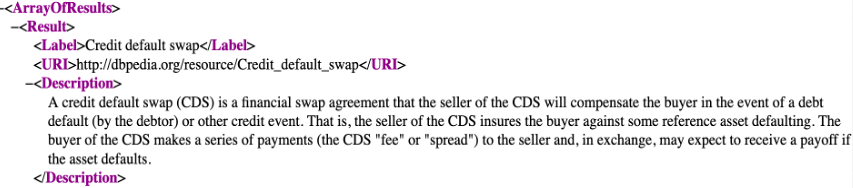}
  \caption{Sample output from DBpedia search API}
  \label{fig:dbpedia}
\end{figure*}

\textbf{Definitions from DBpedia}: We used the DBpedia search API\footnote{https://lookup.dbpedia.org/api/search} to extract the description of the terms present in the train and test sets. We present such an example in Figure \ref{fig:dbpedia}. In addition to the description, the label was also retained from the result payload to identify the right description for the input terms. We tried token overlap-based similarity of input terms with both matching labels and descriptions. We decided to use the label to term match for description matching after going through a randomly drawn sample. We cleaned both input terms and labels from DBpedia results by converting them to lower case, replacing punctuations by space, removing repetitive spaces, and singularizing the text. We calculated the token overlap ratios for cleaned term and DBpedia labels using these formulas:
$Ratio1 = length(s_{1} \cap s_{2})/length(s_{1})$,
$Ratio2 = length(s_{2})/length(s_{1})$ where s\textsubscript{1} and s\textsubscript{2} represents sets of tokenized cleaned terms and tokenized cleaned DBpedia labels respectively. We empirically decided to use all the instances with $Ratio1 = 1$ and $Ratio2 <= 1.25$ for matching a DBpedia label (and hence description) to the input term.\\

\textbf{Definitions from Investopedia and FIBO}: Inspired by \cite{saini-2020-anuj-investopedia}, we obtained definitions of the terms present in Investopedia’s data dictionary\footnote{https://www.investopedia.com/financial-term-dictionary-4769738} by crawling it. We downloaded a glossary of financial terms from the website of FIBO. We cleaned all the terms from the train and test set and also the terms present in Investopedia's data dictionary using the steps described in the above DBpedia section. We then assigned the Investopedia or FIBO definition to the terms from the train and test sets where cleaned terms from train and test data matched to cleaned Investopedia terms perfectly.

The test set which was provided to us had 326 terms. We augmented the original train and test set with the records where we could either find definition or expansion using the above sources. The train set size increased to 1836 records and the test set size increased to 607 after the data augmentation. We present an example of data augmentation for the term ``callable bond" in Table \ref{tab:sampledataaug}. Table \ref{tab:dataaug} states the number of instances we used from each of the sources to augment the data we had.

\begin{table*}[ht]
\centering
\begin{tabular}{|l|l|l|}
\hline
\textbf{Expanded Term/Term Definition} & \textbf{Label} & \textbf{Source}                                                           \\ \hline
Callable bond                          & Bonds          & \begin{tabular}[c]{@{}l@{}}original and \\ acronym expansion\end{tabular} \\ \hline
\begin{tabular}[c]{@{}l@{}}bond that includes a stipulation allowing the issuer \\ the right to repurchase and retire   the bond at the call price after the call protection period\end{tabular} &
  Bonds &
  FIBO \\ \hline
\begin{tabular}[c]{@{}l@{}}A callable bond (also called redeemable bond) is a type of bond (debt security) that allows\\ the issuer of the bond to retain the privilege of redeeming the bond at some   point before \\ the bond reaches its date of maturity.\end{tabular} &
  Bonds &
  DBpedia \\ \hline
\end{tabular}
\caption{Result of Data Augmentation of the term "Callable bond"}
\label{tab:sampledataaug}
\end{table*}

\begin{table}
\centering
\begin{tabular}{|l|l|}
\hline
\textbf{Data Source}    & \textbf{Count} \\ \hline
Original modelling data & 1040           \\ \hline
DBpedia                 & 257            \\ \hline
FIBO                    & 236            \\ \hline
Investopedia            & 85             \\ \hline
Acronym expansion       & 218            \\ \hline
\end{tabular}
\caption{Details of various data sources}
\label{tab:dataaug}
\end{table}

\section{System Description}
We tried to solve this problem as the term classification and term similarity problems. Two of our 3 submissions are modelled as the term classification problem, whereas the third system is designed to be a phrase/sentence similarity problem between terms (or expanded terms from the augmented dataset) and the definitions of 17 class labels that were extracted from FIBO / Internet. All the systems rely on semantic similarity and use FinBERT model to generate the term or token embedding representations.
We divided the given data into training and validation sets having 832 and 208 terms respectively.

\subsection{System - 1 (S1)}
This is the simplest of our proposed systems, where we did not use the augmented dataset and used only the original set that was shared by organizers. We loaded FinBERT pre-trained model and fine-tuned it by trying to classify the representation of [CLS] token into one of the 17 labels mentioned previously. 
Since the original data did not have longer terms, we kept the maximum length to 32, and train and validation batch sizes of 64. We used Adam optimizer with a learning rate of 0.00002. We ran the model for 40 epochs and picked the model saved after 18\textsuperscript{th} epoch based on the performance on the validation set. Finally, we ranked the predictions based on the predicted probability of each class.

\subsection{System - 2 (S2)}
This system is similar to System-1 with the only difference that data being the augmented set and not the original dataset.
Since the augmented dataset had the descriptions of the terms, the inputs were considerably longer. Hence, we increased the maximum length to 256 while keeping all the other hyper-parameters the same. After, training the model for 40 epochs we selected the model saved after the 17\textsuperscript{th} epoch as the best model based on validation set performance.

\subsection{System -3 (S3)} 
\label{sec:s3}
\begin{figure*}
\centering
  \includegraphics[width=10cm,height=5.65cm]{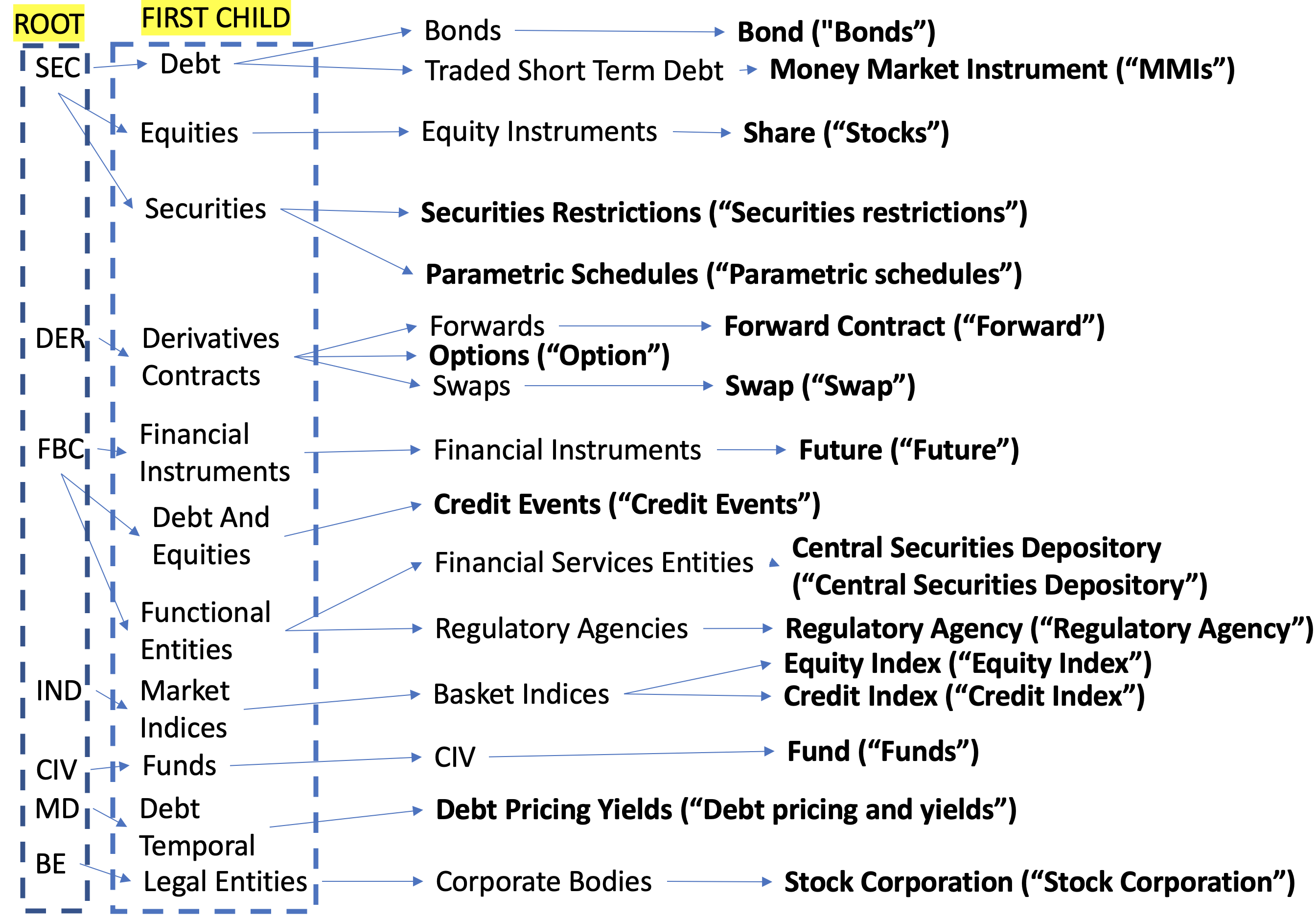}
  \caption{Label Hierarchy from FIBO. Bold (leaf nodes) denotes the labels.}
  \label{fig:hier}
\end{figure*}

We explored the FIBO ontology to understand the hierarchy \cite{jsi-2021-finsim} of the 17 labels as depicted in Figure \ref{fig:hier}. We used the augmented data described in section \ref{sec:dataaug} to create a labelled dataset having similarity scores. For every term definition (T) to label definition (L) mapping which existed in the extended training set, we assigned a similarity score of 1.0 to the (T,L) pair and picked up 10 training instances randomly ensuring none of their label definition was same as L. For each of the label definitions (LL) present in this sample, we extracted its root node and first child node. We did the same for the original label definition (L). Then, we compared these nodes. If the root node and first child node of L were different from that of LL then we assigned a similarity score of 0 to the (T, LL) pair. If the root nodes were the same, we assigned a similarity score of 'k' when the first child nodes differed and a similarity score of '2k' when they were the same (where $0<k<1$). We empirically figured out that k=0.4 works the best. As expected, the number of instances with a similarity score equal to 0 increased substantially. We under-sampled such instances and the new training set had 30\% instances with similarity score 1.0, 12\% instances with similarity score 'k', 28\% instances with similarity score '2k' and 30\% instances with similarity score 0.
After that, we fine-tuned a FinBERT \cite{araci2019finbert} model using Sentence BERT \cite{sbert} framework with this newly generated labelled data for 25 epochs with a batch size of 20. Our objective was to minimize the multiple negatives ranking loss and online contrastive loss. We used a margin of 0.5 and cosine distance as a distance metric while training this model. Finally, we converted all of the 17 labels' definitions and term definitions from the validation set to vectors using this fine-tuned model. For every such term definition, we performed a semantic search over the label vectors and ranked them in decreasing order of cosine similarity.\\
System 2 and 3 take advantage of term expansion during both model training and scoring phases, which causes certain observations to appear more than once (reference: Table \ref{tab:dataaug}). We derive the final prediction by averaging the output probabilities for all the 17 classes for all the occurrences of the term.

\section{Experimentation and Results}
We had 1040 observations after removing the duplicates. We did an 80:20 split to create a training and validation set from this. We augmented the given modelling set by incorporating definitions from DBpedia, FIBO and Investopedia. We used the list of acronyms extracted from the prospectus corpus to create a copy with acronym expansion. This helped us to increase the original data to 1836 records (mentioned in Table \ref{tab:labeldis}). It should be noted that we could not find the expansions for all the terms given in the modelling set. Train and validation set sizes for the original modelling set and expanded data were (832 \& 208) and (1470 \& 366) respectively.

We established a baseline by running the scripts provided by the organizers. Then, we considered original modelling data and fine-tuned base BERT-cased model \cite{devlin-etal-2019-bert} to predict the class label by taking the representation of [CLS] token while passing it through few layers of a feed-forward network. This performed better than baseline. We then tried the same BERT-base model on the expanded dataset, which gave us further performance improvement. Since the only  major change between these runs was the data, the improvement can be attributed to the expanded data.

We experimented with a few of the other pre-trained models that are available on the Huggingface model repository \cite{wolf2020huggingfaces}. We observed clear improvement when we used the FinBERT model which was trained on data specific to the financial domain. The model performance successively increased when we used a combination of data expansion with FinBERT. Furthermore, we tried to fine-tune FinBERT using Sentence Transformers \cite{sbert} to capture semantic textual similarity. For this, we used several combinations of term and term definitions with label and label definitions.

All the hyperparameters for the final 3 models have been already mentioned in the system description. After rigorous experimentation, these hyperparameters were selected empirically based on validation set performance. The results are presented in Table \ref{tab:results}. Since the number of submissions was restricted to 3 for each team, we do not have the performance numbers of the BERT models in the test set. Analysing the results we see that SentenceBERT trained with FinBERT at the backed as mentioned in section-\ref{sec:s3} performed the best.
\begin{table}
\centering
\begin{tabular}{|l|l|l|l|l|l|}
\hline
               &               & \multicolumn{2}{l|}{\textbf{Validation set}} & \multicolumn{2}{l|}{\textbf{Test set}} \\ \hline
\textbf{Model} & \textbf{Data} & \textbf{Rank}    & \textbf{Acc.}    & \textbf{Rank}    & \textbf{Acc.}   \\ \hline
Base-1         & Org.     & 2.158              & 0.498              & 1.941            & 0.564           \\ \hline
Base-2         & Org.      & 1.201              & 0.876              & 1.75             & 0.669           \\ \hline
BERT           & Org.      & 1.177            & 0.899            & -                & -               \\ \hline
BERT           & Ext.     & 1.153            & 0.928            & -                & -               \\ \hline
FinBERT(S1)        & Org.      & 1.117            & 0.928            & 1.257            & 0.886           \\ \hline
FinBERT(S2)       & Ext.      & 1.110            & 0.942            & 1.220            & 0.895           \\ \hline
SBERT(S3)         & Ext.     & \textbf{1.086}   & \textbf{0.947}   & \textbf{1.156}   & \textbf{0.917}  \\ \hline
\end{tabular}
\caption{Results on validation and test set. Org. represents original and Ext. represents extended. Base refers to baseline.}
\label{tab:results}
\end{table}

\section{Conclusion and Future Works}
In this work, we attempted to solve the hypernym and synonym discovery hosted at FinSim-3. This challenge aimed to enable the better use of ontologies like FIBO using hypernyms and synonyms, and we used these ontologies themselves to develop our systems which perform significantly better than the provided baseline systems. This proves the present use of these ontologies. The presented solution is recursive in a sense as it uses knowledge from ontologies to further increase the effectiveness and use of the same. Apart from data augmentation, our solution relies upon semantic similarity learnt from pre-trained embedding models that were learnt on the relevant domain. We observed the clear benefits of domain specific pretraining during the experimentation.

In future, we would like to explore Knowledge Graphs (as described in \cite{portisch2021finmatcher}) to further improve the improve performance of the models. We also want to explore other variants of FinBERT \cite{araci2019finbert} and fine-tune them using the Masked Language Modeling technique (as mentioned by the winner of FinSim-2 \cite{pplyu-2021-finsim}) and Next Sentence Prediction objective. Moreover, this research can be extended by extracting sentences present in the prospectus (similar to \cite{tcs2021finsim}) to create more positive and negative samples.

\bibliographystyle{named}
\bibliography{finsim}

\end{document}